\documentclass[11pt,a4paper]{article}
\usepackage[hyperref]{acl2017}
\usepackage{times, graphicx, soul}
\usepackage{latexsym}
\usepackage{url}
\usepackage{mathptm}

\aclfinalcopy 
\def\aclpaperid{3180} 

\title{Disfluency Detection using a Noisy Channel Model and 
a Deep Neural Language Model}

\author{Paria Jamshid Lou \\
  Department of Computing \\
  Macquarie University \\
  Sydney, Australia\\
  {\tt paria.jamshid-lou@hdr.mq.edu.au} \\\And
  Mark Johnson \\
  Department of Computing \\
  Macquarie University \\    
  Sydney, Australia\\
  {\tt mark.johnson@mq.edu.au} \\}

\date{4/22/2017}

\begin{document}
\maketitle
\begin{abstract}
This paper presents a model for disfluency detection in spontaneous speech transcripts called \emph{LSTM Noisy Channel Model}. The model uses a Noisy Channel Model (NCM) to generate $n$-best candidate disfluency analyses and a Long Short-Term Memory (LSTM) language model to score the underlying fluent sentences of each analysis. The LSTM language model scores, along with other features, are used in a MaxEnt reranker to identify the most plausible analysis. We show that using an LSTM language model in the reranking process of noisy channel disfluency model improves the state-of-the-art in disfluency detection.
\end{abstract}

\section{Introduction}
Disfluency is a characteristic of spontaneous speech which is not present in written text. Disfluencies are informally defined as interruptions in the normal flow of speech that occur in different forms, including false starts, corrections, repetitions and filled pauses. According to Shriberg's~\citeyearpar{shri:94} definition, the basic pattern of speech disfluencies contains three parts: \textit{reparandum\footnote{Reparandum is sometimes called \textit{edit}.}}, \textit{interregnum} and \textit{repair}. Example~\ref{ex:BostonDenver} illustrates a disfluent structure, where the reparandum \textit{to Boston} is the part of the utterance that is replaced, the interregnum \textit{uh, I mean} is an optional part of a disfluent structure that consists of a filled pause \emph{uh} and a discourse marker \emph{I mean} and the repair \textit{to Denver} replaces the reparandum. The fluent version of  Example~\ref{ex:BostonDenver} is obtained by deleting reparandum and interregnum words. 
\begin{equation}
\begin{array}{l}
\mbox{\it I want a flight~}\overbrace{\mbox{\it\strut to Boston,}}^{\mbox{\tiny reparandum}}\\
\strut\hspace{-0.5cm}\underbrace{\mbox{\it{}uh, I mean\strut}}_{\mbox{\tiny interregnum}} \underbrace{\mbox{\it\strut to Denver~}}_{\mbox{\tiny repair}} {\it\strut on~Friday}
\end{array}
\end{equation}
\label{ex:BostonDenver}

While disfluency rate varies with the context, age and gender of speaker,~\citet{bort:01} reported disfluencies once in every 17 words. Such frequency is high enough to reduce the readability of speech transcripts. Moreover, disfluencies pose a major challenge to natural language processing tasks, such as dialogue systems, that rely on speech transcripts~\citep{ost:07}. Since such systems are usually trained on fluent, clean corpora, it is important to apply a speech disfluency detection system as a pre-processor to find and remove disfluencies from input data. By disfluency detection, we usually mean identifying and deleting reparandum words. Filled pauses and discourse markers belong to a closed set of words, so they are trivial to detect~\citep{john:04}.

In this paper, we introduce a new model for detecting restart and repair disfluencies in spontaneous speech transcripts called \emph{LSTM Noisy Channel Model (LSTM-NCM)}. The model uses a Noisy Channel Model (NCM) to generate $n$-best candidate disfluency analyses, and a Long Short-Term Memory (LSTM) language model to rescore the NCM analyses. The language model scores are used as features in a MaxEnt reranker to select the most plausible analysis. We show that this novel approach improves the current state-of-the-art.

\section{Related Work}
\label{ssec:first}
Approaches to disfluency detection task fall into three main categories: sequence tagging, parsing-based and noisy channel model. 
The sequence tagging models label words as fluent or disfluent using different techniques, including conditional random fields~\citep{ost:13, zay:14, fer:15}, hidden Markov models~\citep{liu:06, schu:10} or 
recurrent neural networks~\citep{hou:15, zay:16}. Although sequence tagging models can be easily generalized to a wide range of domains, they require a specific state space for disfluency detection, such as begin-inside-outside (BIO) style states that label words as being inside or outside of a reparandum word sequence. The parsing-based approaches refer to parsers that detect disfluencies, as well as identifying the syntactic structure of the sentence~\citep{ras:13, hon:14, yoshi:16}. Training a parsing-based model requires large annotated tree-banks that contain both disfluencies and syntactic structures. Noisy channel models (NCMs) use the similarity between reparandum and repair as an indicator of disfluency. However, applying an effective language model (LM) inside an NCM is computationally complex. To alleviate this problem, some researchers use more effective LMs to rescore the NCM disfluency analyses. Johnson and Charniak~\shortcite{john:04} applied a syntactic parsing-based LM trained on the fluent version of the Switchboard corpus to rescore the disfluency analyses. Zwarts and Johnson~\shortcite{zwa:11} trained external $n$-gram LMs on a variety of large speech and non-speech corpora to rank the analyses. Using the external LM probabilities as features to the reranker improved the baseline NCM~\citep{john:04}. The idea of applying external language models in the reranking process of the NCM motivates our model in this work. 

\section{LSTM Noisy Channel Model}
\label{ssec:second}
We follow Johnson and Charniak~\shortcite{john:04} in using an NCM to find the $n$-best candidate disfluency analyses for each sentence. The NCM, however, lacks an effective language model to capture more complicated language structures. To overcome this problem, our idea is to use different LSTM language models to score the underlying fluent sentences of the analyses proposed by the NCM and use the language model scores as features to a MaxEnt reranker to select the best analysis. In the following, we describe our model and its components in details.  

In the NCM of speech disfluency, we assume that there is a well-formed source utterance $X$ to which some noise is added and generates a disfluent utterance $Y$ as follows.

\vspace{0.1cm} 	
\begin{center}
X = a flight to Denver\\
Y = a flight \textit{to Boston uh I mean} to Denver
\end{center}
\vspace{0.1cm} 

Given $Y$, the goal of the NCM is to find the most likely source sentence $\hat{X}$ such that:
\vspace{0.1cm}
\begin{equation}\label{eq:01}
\hat X = \arg\max_{X}P(Y|X)P(X)
\end{equation}

As shown in Equation~\ref{eq:01}, the NCM contains two components: the channel model $P(Y|X)$ and the language model $P(X)$. Calculating the channel model and language model probabilities, the NCM generates $25$-best candidate disfluency analyses as follows.
\begin{equation} \label{ex:02}
\centering
\includegraphics[width=0.4\textwidth]{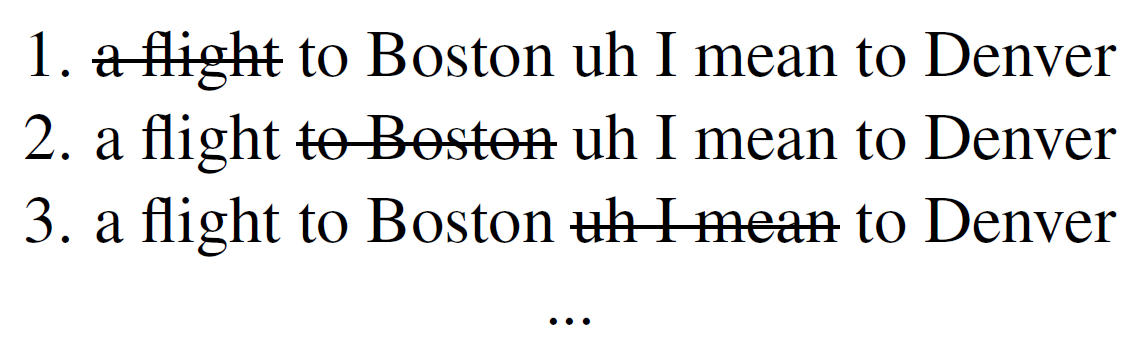}
\end{equation}
Example~\ref{ex:02} shows sample outputs of the NCM, where potential reparandum words are specified with strikethrough text. The MaxEnt reranker is applied on the candidate analyses of the NCM to select the most plausible one. 

\subsection{Channel Model}
\label{sub:one}
We assume that $X$ is a substring of $Y$, so the source sentence $X$ is obtained by deleting words from $Y$. For each sentence $Y$, there are only a finite number of potential source sentences. However, with the increase in the length of $Y$, the number of possible source sentences $X$ grows exponentially, so it is not feasible to do exhaustive search. Moreover, since disfluent utterances may contain an unbounded number of crossed dependencies, a context-free grammar is not suitable for finding the alignments. The crossed dependencies refer to the relation between repair and reparandum words which are usually the same or very similar words in roughly the same order as in Example~\ref{ex:01}.
\begin{equation} \label{ex:01}
\centering
\includegraphics[width=0.35\textwidth]{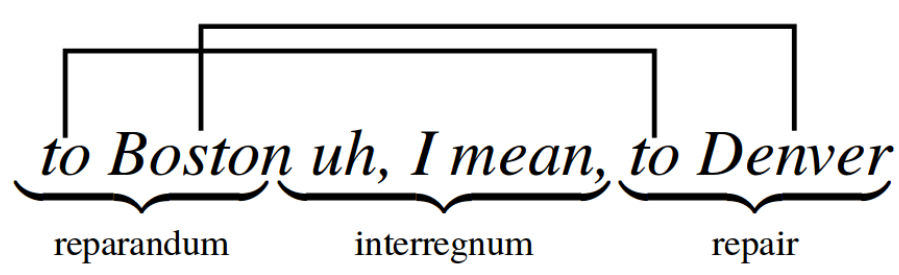}
\end{equation}

We apply a Tree Adjoining Grammar (TAG) based transducer~\citep{john:04} which is a more expressive formalism and provides a systematic way of formalising the channel model. The TAG channel model encodes the crossed dependencies of speech disfluency, rather than reflecting the syntactic structure of the sentence. The TAG transducer is effectively a simple first-order Markov model which generates each word in the reparandum conditioned on the preceding word in the reparandum and the corresponding word in the repair. More details about the TAG channel model can be found in~\cite{john:04}. 

\subsection{Language Model}
\label{sub:two}
The language model of the NCM evaluates the fluency of the sentence with disfluency removed. The language model is expected to assign a very high probability to a fluent sentence $X$ (e.g. \textit{a flight to Denver}) and a lower probability to a sentence $Y$ which still contains disfluency (e.g. \textit{a flight to Boston uh I mean to Denver}). However, it is computationally complex to use an effective language model within the NCM. The reason is the polynomial-time dynamic programming parsing algorithms of TAG can be used to search for likely repairs if they are used with simple language models such as a bigram LM~\citep{john:04}. The bigram LM within the NCM is too simple to capture more complicated language structure. In order to alleviate this problem, we follow Zwarts and Johnson~\shortcite{zwa:11} by training LMs on different corpora, but we apply state-of-the-art recurrent neural network (RNN) language models.

\subsubsection*{LSTM}
\label{sub:three}
We use a long short-term memory (LSTM) neural network for training language models. LSTM is a particular type of recurrent neural networks which has achieved state-of-the-art performance in many tasks including language modelling~\citep{mikolov:10, joz:16}. LSTM is able to learn long dependencies between words, which can be highly beneficial for the speech disfluency detection task. Moreover, it allows for adopting a distributed representation of words by
constructing word embedding~\citep{mikolov:13}.

We train forward and backward (i.e. input sentences are given in reverse order) LSTM language models using truncated backpropagation through time algorithm~\citep{rum:86} with mini-batch size $20$ and total number of epochs $13$. The LSTM model has two layers and $200$ hidden units. The initial learning rate for stochastic gradient optimizer is chosen to $1$ which is decayed by $0.5$ for each epoch after maximum epoch $4$. We limit the maximum sentence length for training our model due to the high computational complexity of longer histories in the LSTM. In our experiments, considering maximum $50$ words for each sentence leads to good results. The size of word embedding is $200$ and it is randomly initialized for all LSTM LMs\footnote{All code is written in TensorFlow~\citep{tens:15}}.

Using each forward and backward LSTM language model, we assign a probability to the underlying fluent parts of each candidate analysis.

\subsection{Reranker}
\label{sub:reran}
In order to rank the the $25$-best candidate disfluency analyses of the NCM and select the most suitable one, we apply the MaxEnt reranker proposed by Johnson et al.~\shortcite{john:04a}. We use the feature set introduced by Zwarts and Johnson~\shortcite{zwa:11}, but instead of $n$-gram scores, we apply the LSTM language model probabilities. The features are so good that the reranker without any external language model is already a state-of-the-art system, providing a very strong baseline for our work. The reranker uses both model-based scores (including NCM scores and LM probabilities) and surface pattern features (which are boolean indicators) as described in Table~\ref{tab:01}. Our reranker optimizes the expected f-score approximation described in Zwarts and Johnson~\shortcite{zwa:11} with L2 regularisation.

\begin{table}[h!] 
\begin{tabular}{|p{7.3cm}|}
\hline \bf model-based features \\ \hline
1-2. forward \& backward LSTM LM scores  \\
3-7. log  probability of the entire NCM \\
8. sum of the log LM probability \& the log channel model probability plus number of edits in the sentence \\
9. channel model probability \\
\hline \bf surface pattern features\\ \hline
10. CopyFlags\_X\_Y: if there is an exact copy in the input text of length $X$ ($1 \leq X \leq 3$) and the gap between the copies is $Y$ ($0 \leq Y \leq 3$) \\
11. WordsFlags\_L\_n\_R: number of flags to the left (L) and to the right (R) of a 3-gram area ($0 \leq L, R \leq 1$) \\
12. SentenceEdgeFlags\_B\_L: it captures the location and length of disfluency. The Boolean B sentence initial or sentence final disfluency, L ($1 \leq L \leq 3$) records the length of the flags.  \\
\hline
\end{tabular}
\caption{The features used in the reranker. They, except for the first and second one, were applied by Zwarts and Johnson~\shortcite{zwa:11}.}
\label{tab:01}
\end{table}

\section{Corpora for Language Modelling}
\label{ssec:third}
In this work, we train forward and backward LSTM language models on Switchboard~\citep{godfrey:93} and Fisher~\citep{cieri:04} corpora. Fisher consists of $2.2\times10^7$ tokens of transcribed text, but disfluencies are not annotated in it. Switchboard is a widely available corpus ($1.2\times10^6$ tokens) where disfluencies are annotated according to Shriberg's~\shortcite{shri:94} scheme. Since the bigram language model of the NCM is trained on this corpus, we cannot directly use Switchboard
to build LSTM LMs. The reason is that if the training data of Switchboard is used both for predicting language fluency and optimizing the loss function, the reranker will overestimate the weight related to the LM features extracted from Switchboard. This is because the fluent sentence itself is part of the language model~\citep{zwa:11}. As a solution, we apply a $k$-fold cross-validation ($k=20$) to train the LSTM language models when using Switchboard corpus. 

We follow Charniak and Johnson~\shortcite{char:01} in splitting Switchboard corpus into training, development and test set. The training data consists of all sw[23]$\ast$.dps files, development training consists of all sw4[5-9]$\ast$.dps files and test data consists of all sw4[0-1]$\ast$.dps files. Following Johnson and Charniak~\shortcite{john:04}, we remove all partial words and punctuation from the training data. Although partial words are very strong indicators of disfluency, standard speech recognizers never produce them in their outputs, so this makes our evaluation both harder and more realistic. 

\section{Results and Discussion}
\label{ssec:forth}
We assess the proposed model for disfluency detection with all MaxEnt features described in Table~\ref{tab:01} against the baseline model. The noisy channel model with exactly the same reranker features except the LSTM LMs forms the baseline model. 

To evaluate our system, we use two metrics \emph{f-score} and \emph{error rate}. Charniak and Johnson~\shortcite{char:01} used the f-score of labelling reparanda or ``edited" words, while Fiscus et al~\shortcite{fis:04} defined an ``error rate" measure, which is the number of words falsely labelled divided by the number of reparanda words. Since only $6\%$ of words are disfluent in Switchboard corpus, accuracy is not a good measure of system performance. F-score, on the other hand, focuses more on detecting ``edited" words, so it is a decent metric for highly skewed data. 

According to Tables~\ref{tab:02} and \ref{tab:03}, the LSTM noisy channel model outperforms the baseline. The experiment on Switchboard and Fisher corpora demonstrates that the LSTM LMs provide information about the global fluency of an analysis that the local features of the reranker do not capture. The LSTM language model trained on Switchboard corpus results in the greatest improvement. Switchboard is in the same domain as the test data and it is also disfluency annotated. Either or both of these might be the reason why Switchboard seems to be better in comparison with Fisher which is a larger corpus and might be expected to make a better language model. Moreover, the backward LSTMs have better performance in comparison with the forward ones. It seems when sentences are fed in reverse order, the model can more easily detect the unexpected word order associated with the reparandum to detect disfluencies. In other words, that the disfluency is observed ``after" the fluent repair in a backward language model is helpful for recognizing disfluencies. 

\vspace{0.2cm} 
\begin{table}[h!]
\begin{center}
\begin{tabular}{|l|c|c|c|}
\hline baseline & \multicolumn{3}{|c|}{$85.3$} \\ 
\hline \bf corpus & \bf forward & \bf backward & \bf both \\ \hline
Switchboard & $86.1$  & $86.6$ & $86.8$  \\  \hline
Fisher & $86.2$ & $86.5$ & $86.3$\\ \hline
\end{tabular}
\end{center}
\caption{F-scores on the dev set for a variety of LSTM language models.}\label{tab:02} 
\end{table}
\vspace{0.3cm} 
\begin{table}[h!]
\begin{center}

\begin{tabular}{|l|c|c|c|}
\hline baseline & \multicolumn{3}{|c|}{$27.0$} \\ 
\hline \bf corpus & \bf forward & \bf backward & \bf both \\ \hline
Switchboard & $25.5$  & $24.8$ & $24.3$\\  \hline
Fisher & $25.6$ & $25.0$ & $25.3$ \\ \hline
\end{tabular}
\end{center}
\caption{Expected error rates on the dev set for a variey of LSTM language models.}\label{tab:03} 
\end{table}
\vspace{0.3cm}

We compare the performance of Kneser-Ney smoothed $4$-gram language models with the LSTM corresponding on the reranking process of the noisy channel model. We estimate the $4$-gram models and assign probabilities to the fluent parts of disflueny analyses using the SRILM toolkit~\citep{sto:02}. As Tables~\ref{tab:04} and \ref{tab:05} show including scores from a conventional $4$-gram language model does not improve the model's ability to find disfluencies, suggesting that the LSTM model contains all the useful information that the $4$-gram model does. In order to give a more general idea on the performance of LSTM over standard LM, we evaluate our model when the language model scores are used as the only features of the reranker. The f-score for the NCM alone without applying the reranker is $78.7$, while using $4$-gram language model scores in the reranker increases the f-score to $81.0$. Replacing the $4$-gram scores with LSTM language model probabilities leads to further improvement, resulting an f-score $82.3$.

\begin{table}[h!]
\begin{center} 
\begin{tabular}{|l|c|c|c|}
\hline baseline & \multicolumn{3}{|c|}{$85.3$} \\ 
\hline \bf corpus & \bf $4$-gram & \bf LSTM & \bf both \\ \hline
Switchboard & $85.1$ & $86.8$ & $86.1$ \\  \hline
Fisher & $85.6$ & $86.3$ & $86$ \\ \hline
\end{tabular}
\end{center}
\caption{F-score for $4$-gram, LSTM and combination of both language models.}\label{tab:04} 
\end{table} 
 
\begin{table}[h!]
\begin{center} 
\begin{tabular}{|l|c|c|c|}
\hline baseline & \multicolumn{3}{|c|}{$27.0$} \\ 
\hline \bf corpus & \bf $4$-gram & \bf LSTM & \bf both \\ \hline
Switchboard  & $27.5$ & $24.3$ & $26$ \\  \hline
Fisher & $26.6$ & $25.3$ &  $26$ \\ \hline
\end{tabular}
\end{center}
\caption{Expected error rates for $4$-gram, LSTM and combination of both language models.}\label{tab:05} 
\end{table} 

We also compare our best model on the development set to the state-of-the-art methods in the literature. As shown in Table~\ref{tab:06}, the LSTM noisy channel model outperforms the results of prior work, achieving a state-of-the-art performance of $86.8$. It also has better performance in comparison with Ferguson et al.~\shortcite{fer:15} and Zayat et al.'s~\shortcite{zay:16} models, even though they use richer input that includes prosodic features or partial words. 
\vspace{0.2cm} 

\begin{table}[h!]
\begin{center}
\begin{tabular}{|l|c|}
\hline \bf Model & \bf f-score  \\ \hline
\citet{yoshi:16} & $62.5$ \\
\citet{john:04} & $79.7$ \\
\citet{john:04a} & $81.0$   \\
\citet{ras:13} & $81.4$ \\
\citet{qian:13} & $82.1$ \\
\citet{hon:14} & $84.1$ \\
\citet{fer:15} \bf* & $85.4$ \\
\citet{zwa:11} & $85.7$ \\ 
\citet{zay:16} \bf* & $85.9$ \\ \hline
\bf LSTM-NCM &  $\bf 86.8$ \\

\hline
\end{tabular}
\end{center}
\caption{Comparison of the LSTM-NCM to state-of-the-art methods on the dev set. {\bf*}Models have used richer input.}\label{tab:06} 
\end{table}

\section{Conclusion and Future Work}
\label{ssec:fifth}
In this paper, we present a new model for disfluency detection from spontaneous speech transcripts. It uses a long short-term memory neural network language model to rescore the candidate disfluency analyses produced by a noisy channel model. The LSTM language model scores as features in a MaxEnt reranker improves the model's ability to detect and correct restart and repair disfluencies. The model outperforms other models reported in the literature, including models that exploit richer information from the input. As future work, we apply more complex LSTM language models such as sequence-to-sequence on the reranking process of the noisy channel model. We also intend to investigate the effect of integrating LSTM language models into other kinds of disfluency detection models, such as sequence labelling and parsing-based models.

\section*{Acknowledgements}
We would like to thank the anonymous reviewers for their insightful comments and suggestions. 

\bibliography{acl2017}
\bibliographystyle{acl_natbib}

\appendix

\end{document}